\documentclass{article}
\usepackage{spconf,amsmath,graphicx}

\usepackage{multirow}
\usepackage[]{bm}
\usepackage{booktabs}
\usepackage{rotating}
\usepackage{amsfonts}
\usepackage{graphicx,wrapfig}
\usepackage{subcaption}
\usepackage{mathtools}
\usepackage{amsmath,amssymb,amsfonts}
\usepackage{color,soul}
\usepackage[bookmarks=false]{hyperref}

\title{Distributional Shifts in Automated Diabetic Retinopathy Screening}
\name{Jay Nandy$^1$~~~~~~~~~~~Wynne Hsu$^{1,2}$~~~~~~~~~~~Mong Li Lee$^{1,2}$}
\address{$^1$School of Computing, National University of Singapore \\
	$^2$Institute of Data Science, National University of Singapore \\
	 \{jaynandy,whsu,leeml\}@comp.nus.edu.sg
}
\begin{document}
%
\maketitle
\begin{abstract}
Deep learning-based models are developed to automatically detect if a retina image is `referable' in diabetic retinopathy (DR) screening. 
However, their classification accuracy degrades as the input images distributionally shift from their training distribution. 
Further, even if the input is not a retina image, a standard DR classifier produces a high confident prediction that the image is `referable'.
Our paper presents a Dirichlet Prior Network-based framework to address this issue. 
It utilizes an out-of-distribution (OOD) detector model and a DR classification model to improve generalizability by identifying OOD images.
Experiments on real-world datasets indicate that the proposed framework can eliminate the unknown non-retina images and identify the distributionally shifted retina images for human intervention.
\end{abstract}

\section{Introduction}
\label{sec:intro}

Diabetic retinopathy (DR) is one of the leading causes of preventable blindness in the world.
It affects diabetic patients within the first two decades of the disease \cite{drReview_2016}.
Vision loss due to diabetic retinopathy is irreversible. 
Several frameworks are proposed to automate the DR screening process \cite{rdr_ictai_2016,svm_retina_2018}.
Recently, deep neural network (DNN) based models achieve clinically acceptable classification accuracy to detect referable DR at lower costs \cite{rdr_jama_2017,redLesion_2018}.
However, these DNN models are sensitive to \textit{in-domain} training distribution \cite{advStart_iclr_2014,fgsm_iclr_2015,rbfcnn_ijcnn_2020,common_iclr_2019,adaptBN_cp_nips_2020,robustBN_arxiv_2020}.
Any minor distributional shift leads to over-confident predictions even if they are wrong, producing poor classification performance \cite{data_shift_2009, oe_iclr_2019}.
Hence, predictive uncertainty estimation has emerged as a crucial research direction to inform about possible wrong predictions, thus instilling user's trust in deep learning systems \cite{gal_thesis_2016, ensemble_nips_2017, zhanwei-trust}. 

Predictive uncertainty in a classification model can arise from three sources: model uncertainty, data uncertainty, and knowledge uncertainty \cite{gal_thesis_2016,data_shift_2009}.
\textit{Model uncertainty} captures the uncertainty in estimating the model parameters, conditioning on training data \cite{gal_thesis_2016}.
\textit{Data uncertainty} arises from the natural complexities of the underlying distribution, such as class overlap, label noise, and others \cite{gal_thesis_2016}.
\textit{Knowledge (or distributional) uncertainty} arises due to the \textit{distributional shifts} between the training and test examples, i.e., the test data is \textit{out-of-distribution (OOD)} \cite{data_shift_2009,baseline_iclr_2017}.
For real-world applications, the ability to detect OOD examples can allow manual intervention in an informed way.

\begin{figure}[h]
	\centering
	\begin{subfigure}[t]{0.45\linewidth}
		\centering
		\includegraphics[height=70pt]{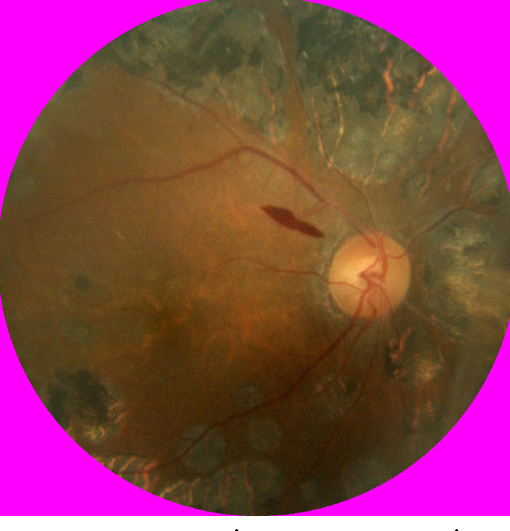}
		\caption{In-domain}
	\end{subfigure}\hspace*{0.2in}
	\begin{subfigure}[t]{0.45\linewidth}
		\centering    
		\includegraphics[height=70pt]{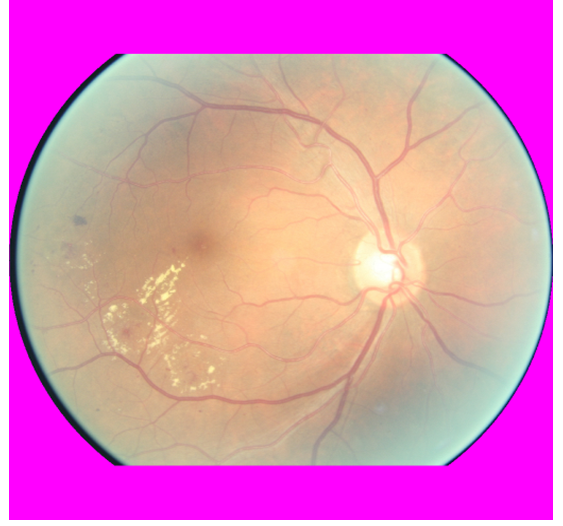}
		\caption{Out-of-distribution}
	\end{subfigure}
	\vspace*{-0.5em}
	\caption{Illustration of the retina images from different sources.}	
	\label{fig:ood_problem}
\end{figure}

To build an automated DR screening system, we typically train a deep learning model using a set of pre-collected retina images  \cite{rdr_jama_2017}.
We apply standard preprocessing techniques (e.g., image normalization and data augmentation) to improve their generalization for unknown test images obtained from the same distribution as the training images.
However, these techniques do not generalize a model for the test images that are distributionally different from those pre-collected training images. 
Figure \ref{fig:ood_problem} illustrates two retina images, obtained from two different distributions.
Hence, a DR classification model may produce incorrect predictions with high confidence for \textit{unknown OOD} images obtained from different distributions.

Recent works have made significant progress to detect distributional uncertainty for unknown OOD test images \cite{baseline_iclr_2017,ensemble_nips_2017,oe_iclr_2019,nandy_dpn_2020}. 
However, these models often fail to detect the OOD examples as the out-distribution and in-distribution become ``alike".
For example, both in-domain and OOD examples are retinal images, as shown in Figure \ref{fig:ood_problem}.
It leads to degrading the performance of these OOD detection models.

In this paper, we focus on the DR screening application. 
We aim to quantify the \textit{distributional shift} in an input retina image while maintaining the high classification performance. 
Our framework utilizes the state-of-the-art Dirichlet prior network (DPN) \cite{dpn2_nips_2019,nandy_dpn_2020}. 
We train an OOD detector separately from the DR classification model.
We use retina images as in-domain and natural images as OOD training set for our DR classifier. 
It also improves their classification performance compared to the baseline CNN model. 
However, it cannot distinguish the out-of-distribution retina images.
Hence, we train a separate OOD detector. 
Here we use both in-domain retina images and OOD images comprising a natural dataset and a few retina images obtained from a different distribution.

Experimental results on multiple real-world datasets demonstrate that our proposed framework effectively detects the OOD retina and non-retina OOD images.
We discard the non-retina images and forward the OOD retina images to the human graders for verification.
Hence, it leads to a greater acceptance of deep learning models for DR screening tasks.

\section{Dirichlet Prior Network}
\label{sec:dpn}

\begin{figure}[htbp]
	\centering
	\begin{subfigure}[t]{0.3\linewidth}
		\centering
		\includegraphics[width=0.6\linewidth]{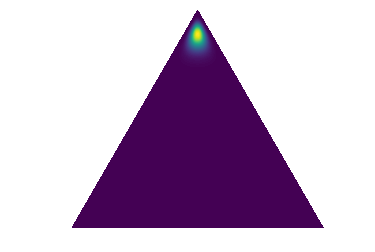}
		\caption{Confident}
		\label{fig:dirichlet_2(a)}
	\end{subfigure}
	\begin{subfigure}[t]{0.3\linewidth}
		\centering    
		\includegraphics[width=0.6\linewidth]{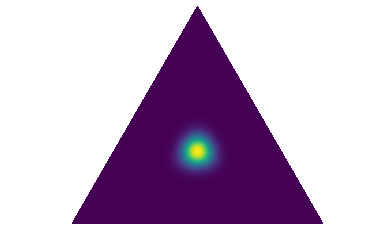}
		\caption{Misclassification}
		\label{fig:dirichlet_2(b)}
	\end{subfigure}
	\begin{subfigure}[t]{0.3\linewidth}
		\centering
		\includegraphics[width=0.6\linewidth]{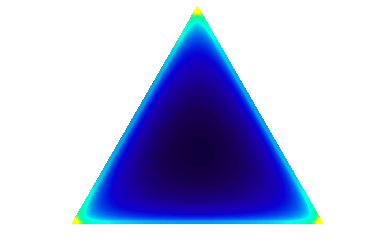}
		\caption{Distributional}
		\label{fig:dirichlet_2(c)}
	\end{subfigure}
\vspace*{-0.5em}
	\caption{Desired output of a DPN classifier.}
	\label{fig:dirichlet_2}

\end{figure}

A Dirichlet Prior Network (DPN) trains a standard neural network with a different loss function to represent their predictions as Dirichlet distributions over the probability simplex \cite{dpn2_nips_2019,nandy_dpn_2020}.
It attempts to produce a sharp Dirichlet at one corner of the simplex when it confidently predicts an in-domain example (see Figure \ref{fig:dirichlet_2}(a)).
For in-domain examples tending to misclassification, it should appear as a sharp distribution in the middle of the simplex, as shown in Figure \ref{fig:dirichlet_2}(b).
For an OOD example, a DPN attempts to produce a sharp multi-modal Dirichlet, spread uniformly at each corner of the simplex to indicate their high \textit{distributional uncertainty} (see Figure \ref{fig:dirichlet_2}(c)) \cite{nandy_dpn_2020,nandy_thesis_2020}.
We observe that the probability densities for Dirichlet distribution in Figure \ref{fig:dirichlet_2}(c) are more scattered over the simplex compared to that in Figures \ref{fig:dirichlet_2}(a) and \ref{fig:dirichlet_2}(b).

A Dirichlet distribution is parameterized with a vector of \textit{concentration parameters} $\bm{\alpha}= {\{\alpha_1, \cdots, \alpha_K\}}$, as follows:
\begin{equation} \small
Dir({\bm \mu | \bm \alpha}) = \frac{\Gamma(\alpha_0)}{\prod_{k=1}^{K}\Gamma(\alpha_k)}\prod_{k=1}^{K} \mu_k^{\alpha_k-1}, ~~ 
\alpha_k >0,
\end{equation}
where $\alpha_0 = \sum_{k=1}^K \alpha_k$ is the \textit{precision} of the distribution.

A higher precision value leads to a sharper uni-modal Dirichlet distribution. 
Consequently, a lower precision produces a flatter uni-modal distribution.
However, as we further uniformly decrease the concentration parameters to lower than $1$, we obtain a \textit{sharp multi-modal distribution} with equal probability density at each corner of the simplex (Figure \ref{fig:dirichlet_2}(c)).
Hence, for a $K$-class classification problem, we need to produce $K$ positive values for each class to obtain the $K$-dimensional Dirichlet distribution.

A deep neural network (DNN) can be viewed as a DPN whose pre-softmax (logit) output corresponding to the class $k$ for an input ${\bm x}$ is $z_k(\bm x)$. 
Then its concentration parameters $\alpha_k$ is given by: $\alpha_k = e^{z_k({\bm x})} $.
The expected posterior for class label $\omega_k$ is given as:
$ p(y = \omega_k | {\bm x}; \bm \theta) 
= \frac{\alpha_k}{\alpha_0}
= \frac{e^{z_k({\bm x})}}{\sum_{k=1}^{K} e^{z_k({\bm x})}}$;
where $\bm \theta$ denotes the DNN parameters.

A DPN measures the \textit{distributional uncertainty} using the \textit{mutual information (MI)} \cite{dpn2_nips_2019}, as follows:
\begin{equation} \small
\vspace{-0.5em}
\label{eq:mi}
\sum_{k=1}^{K} \frac{\alpha_k}{\alpha_0}\big[ \psi(\alpha_k+1)-\psi(\alpha_0+1) -\ln \frac{\alpha_k}{\alpha_0}  \big]
\vspace{-0.5em}
\end{equation}
where $\psi(.)$ is digamma function.
$\alpha_k$ is the concentration parameters for class $k$.
$\alpha_0 = \sum_{k=1}^{K} \alpha_k $ is the precision of the output Dirichlet distributions.
For a \textit{known} in-domain image, a DPN produces a \textit{lower} MI score to indicate low distributional uncertainty.
Consequently, it produces a higher MI score for an OOD image.

\section{Proposed Framework}
\label{sec:method}
\begin{figure}[ht!]\small
	\center{\includegraphics[width=1\linewidth]{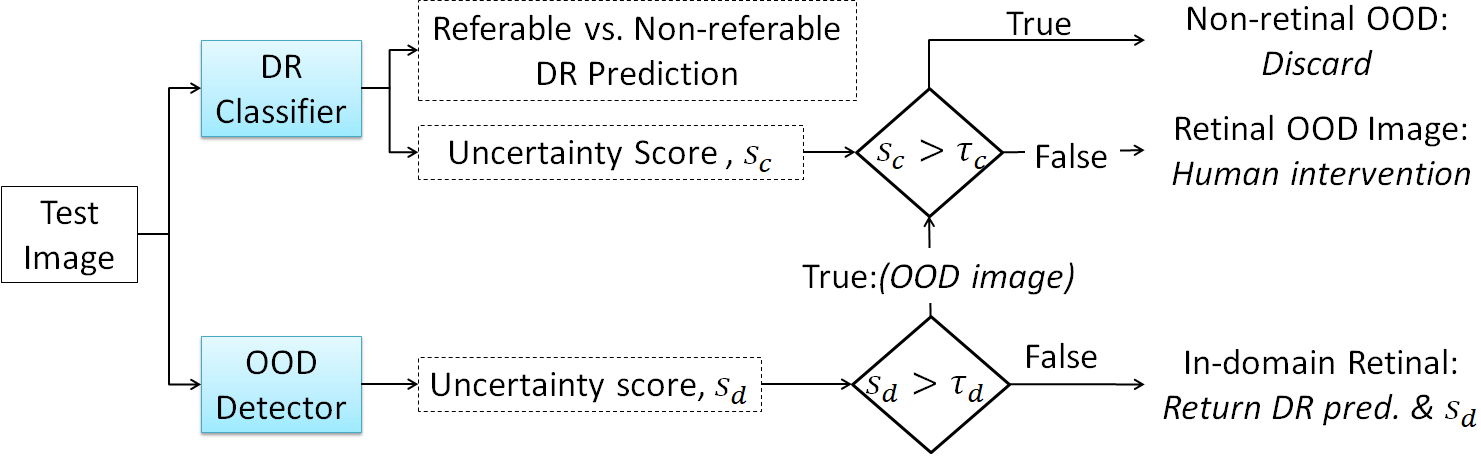}}
	\caption{Overview of our proposed framework.}
	\label{fig:flow}
\end{figure}

Our proposed DPN-based framework for diabetic retinopathy screening utilizes a DR classifier and an OOD detector.
We train the OOD detector separately from the classifier.
Fig. \ref{fig:flow} presents an overview of our proposed framework.
Given an input image, we pass it to both the OOD detector and the DR classifier. 
These two networks produce two different Dirichlet distributions. 
We use  Eq. \ref{eq:mi} to compute the MI scores.  
We denote the scores as $s_d$ and $s_c$ respectively for the  Dirichlet distributions from the OOD detector and DR classifier.
The DR classifier produces lower $s_c$ scores for retina images and higher scores for unknown, non-retina images.
We select a threshold, $\tau_c$, and discard the images with $s_c > \tau_c$ as they are unlikely to be a retina image.
For the OOD detector, we choose another threshold,  $\tau_d$. 
If $s_d<\tau_d$, we accept the input sample is an in-domain retina image.
Hence, if $s_d<\tau_d$ and  $s_c < \tau_c$, we consider the input image is obtained from known in-domain distribution. 
Hence, we can trust the classification prediction without further manual intervention.
Consequently, if $s_d>\tau_d$ and $s_c<\tau_c$, the input is an OOD retina image, and requires human intervention.


\noindent
\textbf{DR Classifier. }
We train a DR classifier using a natural image dataset, $\mathcal{D}_{ood}$ as the OOD training set, along with the original in-domain retina image training set, $\mathcal{D}_{in}$.
The loss function for the DR classifier separately formulates the mean and the precision of the output Dirichlet distributions using the standard cross-entropy loss along with an additional regularization term \cite{nandy_dpn_2020}.
For in-domain training examples $\{\bm x, y\}$, the loss function is given as follows:
\begin{equation}
\small
\label{eq:l_in_reg}
\mathcal{L}_{in}({\bm \theta}; \lambda_{in}) = - \log p({y} | {\bm x}, {\bm \theta}) - \frac{\lambda_{in}}{K}\sum_{c=1}^K\text{sigmoid}(z_c({\bm x})) 
\end{equation}

\noindent
For OOD training examples, the loss function is given as:
\begin{equation}
\small
\label{eq:l_out_reg}
\mathcal{L}_{out}({\bm \theta}; \lambda_{out}) =
\mathcal{H}_{ce} ( \mathcal{U}; p( {y} | {\bm x}, {\bm \theta}))  - \frac{\lambda_{out}}{K}\sum_{c=1}^K\text{sigmoid}(z_c({\bm x})) 
\end{equation}
where $\mathcal{H}_{ce}$ denotes the standard cross-entropy loss.
$\mathcal{U}$ is the uniform distribution over the class labels.

Our DR classifier is trained in a multi-task fashion with the overall loss as:
$\min_{\bm \theta} \mathcal{L}_{in}({\bm \theta}; \lambda_{in}) + \gamma \mathcal{L}_{out}({\bm \theta}; \lambda_{out})$;
where, $\gamma>0$ balances between the in-domain examples and OOD examples.
$\lambda_{in}$ and $\lambda_{out}$ respectively are user-defined hyper-parameters to control the sharpness of the output Dirichlet distributions for in-domain and OOD examples.

The choice of $\lambda_{in} > 0$ produces larger concentration values for in-domain retina images, leading to sharp uni-modal Dirichlet distributions (Figure \ref{fig:dirichlet_2(a)} and Figure \ref{fig:dirichlet_2(b)}).
Consequently, $\lambda_{out} < 0$ enforces the network to produce multi-modal Dirichlet distributions for OOD examples to indicate their high distributional uncertainty (Figure \ref{fig:dirichlet_2(c)}).

\noindent
\textbf{OOD Detector.}
We train the OOD detector using the original in-domain retina images $\mathcal{D}_{in}$, and two OOD datasets, i.e., a natural image dataset, $\mathcal{D}_n$ and a small set of retina images, $\mathcal{D}_r$, obtained from a different source from $\mathcal{D}_{in}$.
We train the OOD detector in a multi-task fashion as follows:
$\min_{\theta} \mathcal{L}_{in}({\bm \theta}; \lambda_{in})  + \gamma_n \mathcal{L}_{n}({\bm \theta}; \lambda_{n}) + \gamma_r \mathcal{L}_{r}({\bm \theta}; \lambda_{r}) $.

Here, $\mathcal{L}_{in}({\bm \theta}; \lambda_{in})$ is corresponding to the in-domain retina training examples, as defined in Equation~\ref{eq:l_in_reg}.
$\mathcal{L}_{n}({\bm \theta}; \lambda_{n})$ and $\mathcal{L}_{r}({\bm \theta}; \lambda_{r})$ are loss functions for $\mathcal{D}_n$ and $\mathcal{D}_r$ respectively, similar to  Equation~\ref{eq:l_out_reg}.
$\gamma_n,~\gamma_r ~>~0$ balance between the loss values for in-domain and different OOD training examples to learn the network parameters ${\bm \theta}$,
$\lambda_{in}, \lambda_{n}$ and $\lambda_{r}$ respectively control the spread of probability mass for the output Dirichlet distributions for the in-domain and the two OOD datasets.
We choose $\lambda_{in} > 0$ to produce sharp uni-modal Dirichlet distributions for in-domain examples, and $\lambda_{n}, \lambda_{n}<0$ to produce multi-modal Dirichlet with uniformly densities at each corner of the simplex for the OOD examples.

\section{Performance Study}
\label{sec:exp}
We evaluate the effectiveness of our  framework for the referable DR screening task using a wide range of datasets:
\begin{itemize}
\item \textit{Kaggle} \cite{dbKaggle}. 
This is a public dataset with 35,126 retina images \cite{kaggle_2}.
We split the dataset into training and test set.
The training set consists of $26,408$ images with $5,129$ referable DR images.
We select a small subset of $1,200$ images from this to train the OOD detector model, denoted as Kaggle-1200.
The test set, Kaggle-Test, has $6,898$ images with $1,354$ referable cases.

\item  \textit{Messidor} \cite{db_messidor}. 
This publicly available dataset has $1200$ retina images, with $501$ referable DR images.

\item \textit{Mayuri}. 
It is a private dataset with $1,520$ retina images with $213$ referable DR images. 

\item \textit{SiDRP}. 
It is a private dataset consisting of retina images from the Singapore National Diabetic Retinopathy Screening Program between 2010-2013.
Our training set, SiDRP-Train, has $89,413$ images with $5,844$ referable DR images, while SiDRP-Test has  $2,239$ images with $1,442$ referable cases.
	
\item  
\textit{ImageNet-Small}.
This is a subset of $25,000$ natural images, randomly selected from ImageNet dataset to train our OOD detector \cite{db_imagenet}.

\item \textit{Non-retina datasets.} We also use \textit{STL10} \cite{db_stl10},  \textit{LSUN} \cite{db_lsun}, \textit{Texture} \cite{db_textures} for our evaluations.
\end{itemize}

\noindent
\textbf{Setup.}
We use VGG-19 \cite{vgg16} for both DR classifier and OOD detector.
We compare the proposed framework with a VGG-19 classifier, denoted as \textit{Baseline}. 
The \textit{Baseline} is trained with cross-entropy loss using the SiDRP-Train dataset.
We train the DR classifier using the in-domain SiDRP-Train and ImageNet-Small as the OOD training set. 
We set the hyper-parameters as $\gamma = 0.1$, $\lambda_{in} = 0.1$ and $\lambda_{out} = -1.0$.
For the OOD detector, we use the in-domain SiDRP-Train and both ImageNet-Small and Kaggle-1200 as OOD training sets.
The hyper-parameters of our OOD detector are set as $\gamma = 0.5$, $\lambda_{in} = 0.5$, $\lambda_{r} = -0.2$ and $\lambda_{n} = -1.0$.
We select the hyper-parameters using validation during training.
\footnote{\small Code modified from \href{https://github.com/jayjaynandy/maximize-representation-gap}{https://github.com/jayjaynandy/maximize-representation-gap}.}

We initialize the model parameters using the pre-trained weights for Imagenet classification task \cite{db_imagenet} as it improves the generalizability of the models \cite{pretrain_icml_2019}.
We re-size the input images to $256 \times 256$ and normalized them using a $5\times5$ median filter to reduce the inconsistency between in-domain training and test images.

\smallskip
\noindent
\textbf{Classification Results under Distributional Shift.}
We first present the performance of our DR Classifier on different test sets.
Table \ref{table:performance} shows the AUROC scores for the referable DR screening task.
We see that both Baseline and  DR Classifier achieve 92.9\% AUROC scores on the in-domain SiDRP-test set.
In contrast,  the performances of both classifiers drop for other DR test sets, confirming the \textit{distributional shifts} of these datasets from the original training set. 
Nevertheless, our proposed DR Classifier leans to produce richer feature representations by incorporating ImageNet-Small for training in an unsupervised fashion.
Hence, it outperforms the Baseline model for these other DR test sets.
\begin{table}[htbp]
	\small
	\centering
	\begin{tabular}{l|c|c}
		\hline
		& Baseline	& DR classifier \\ \hline
		Kaggle-Test &  81.8 &  \textbf{83.7} \\ 
		Messidor &   88.3 &  \textbf{91.0} \\ 
		Mayuri  & 85.6 & \textbf{87.7} \\	
			SiDRP-Test &  \textbf{92.9} & \textbf{92.9} \\ \hline	
	\end{tabular}	
	\vspace{-0.5em}
	\caption{AUROC scores of RDR screening models.}
	\label{table:performance}
\end{table}

\noindent
\textbf{OOD detection performance.}
\label{sec:outlier}
Next, we present the OOD detection performance for unknown natural image datasets and retina datasets obtained from different sources. 
For each image, we compute $s_d$ from the OOD Detector (Equation \ref{eq:mi}). 
We cannot define MI scores for Baseline \cite{dpn2_nips_2019,nandy_thesis_2020}.
Hence, we use \textit{entropy} as their uncertainty score \cite{ensemble_nips_2017}.
We report the percentage of images detected as OOD from the various datasets as we select different thresholds, $\tau_d$.
We obtain these thresholds by dropping $5\%$, $7\%$, and $10\%$ of the in-domain SiDRP-Test images with the top-most uncertainty scores.

Table \ref{table:detection}(a) shows the results for non-retina images.
We can see that the Baseline is unable to distinguish the non-retina images from in-domain retina images.
In contrast, our OOD  detector successfully distinguishes almost all non-retina images even at a $5\%$ threshold.

Table \ref{table:detection}(b) presents the results for OOD retina images.
By incorporating only $1200$ images from Kaggle-Train for training, our OOD detector distinguishes most of the retina images under \textit{distributional shift} in Kaggle-Test as OOD.
For Messidor and Mayuri datasets, our OOD detector significantly outperforms the Baseline by ~20\% on average.

\begin{figure}[htb!]
	\centering

		\centering    
\small	(a) Kaggle-Test \\
		\includegraphics[width=0.8\linewidth]{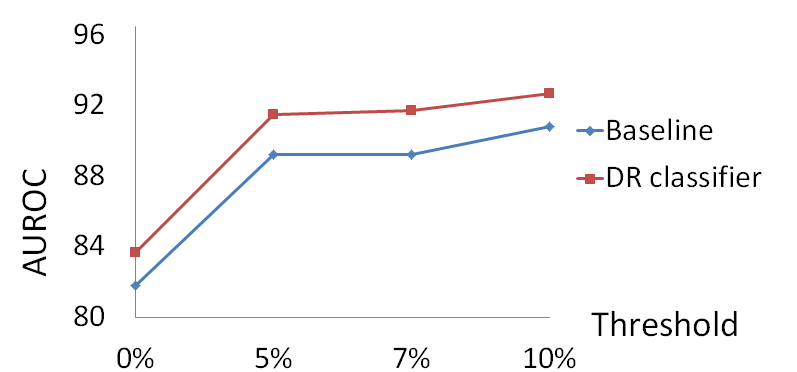}
		\label{fig:kaggle_line}
\\
		\small
		\centering
			(b) Messidor \\
		\includegraphics[width=0.8\linewidth]{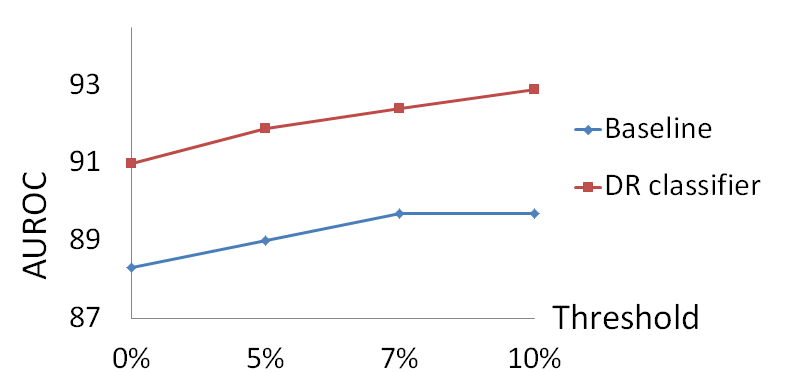}
		\label{fig:messidor_line}
\\
		\small
		\centering
		(c) Mayuri \\
		\includegraphics[width=0.8\linewidth]{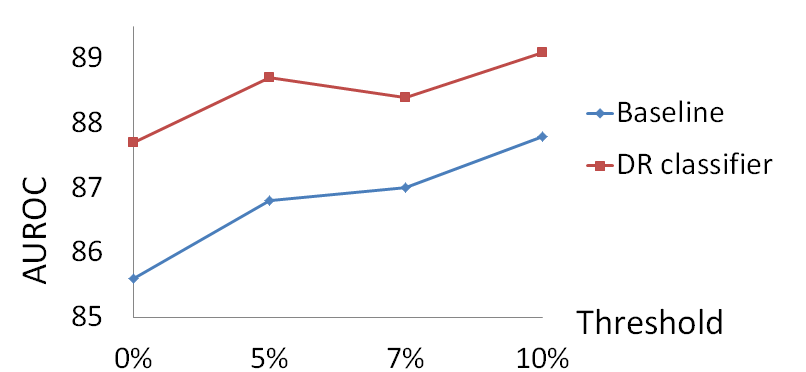}
		\label{fig:mayuri_line}
\vspace{-0.5em}
\caption{AUROC scores after discarding OOD retina images.}
\vspace{-1.0em}
\label{table:final}	
\end{figure}

\smallskip
\noindent
\textbf{Performance after discarding OOD images.}
The objective of our proposed framework is to detect the unknown OOD retina images to improve the trustworthiness of the referable DR screening.
Hence, the overall classification performance should improve after discarding the OOD images.
In our experiment for OOD detection,  we obtain the uncertainty thresholds, $\tau_d$ by discarding $5\%$, $7\%$, and $10\%$ of the in-domain SiDRP-Test images with top-most uncertainty scores.
For the remaining images, we get the predictions from the DR classifier.
Figure \ref{table:final} shows the AUROC scores for referable DR as we increase the threshold to discard the required percentage of OOD images.
We see that the performances of both classifiers improve, with our DR classifier outperforming the Baseline.

\begin{table}[t!]
	\small
		\centering
		(a) Non-Retina Image Datasets \\
		\resizebox{6.5cm}{!}{%
			\begin{tabular}{l|c|c|c}
				\hline
				Dataset	& Threshold & OOD Baseline & OOD Detector \\ \hline
				\multirow{3}{35pt}{STL10} & 5\%	& 0.3	& \textbf{100} \\
				& 7\%& 0.4 & \textbf{100} \\
				& 10\%& 0.5 & \textbf{100} \\ \hline
				
				\multirow{3}{25pt}{LSUN} &	5\%	& 1.2 &	\textbf{100} \\ 
				& 7\%& 1.4 & \textbf{100} \\
				& 10\%& 1.7 & \textbf{100} \\ \hline
				
				\multirow{3}{25pt}{Texture} & 5\% &	1.7	& \textbf{97.7} \\ 
				& 7\%& 1.9 &  \textbf{97.7} \\
				& 10\%& 2.6 & \textbf{97.8} \\ \hline
			\end{tabular}%
		}
		\\
	\vspace{1.0em}
		\centering
		(b) Retina Image Datasets
		\resizebox{6.5cm}{!}{%
			\begin{tabular}{l|c|c|c}
				\hline
				Dataset	& Threshold & OOD Baseline & OOD Detector \\ \hline
				\multirow{3}{35pt}{Kaggle-Test}	& 5\%	& 1.2	& \textbf{90.9} \\
				& 7\%& 1.7 &  \textbf{92.1} \\
				& 10\%& 2.2 & \textbf{93.5} \\ \hline
				
				\multirow{3}{25pt}{Messidor} & 5\%	& 2.3 &	\textbf{17.1} \\
				& 7\%& 2.6 &  \textbf{22.3} \\
				& 10\%& 3.6 & \textbf{30.8} \\ \hline
				\multirow{3}{25pt}{Mayuri} 	&	5\%	&	6.1	& \textbf{21.4} \\
				& 7\%& 7.9 &  \textbf{26.8} \\
				& 10\%& 10.9 & \textbf{34.7} \\ \hline		
			\end{tabular}%
	}
	\vspace{-0.5em}
	\caption{Percentage of OOD images detected.}
	\label{table:detection}
\end{table}

\section{Conclusion}

The ability to distinguish unknown OOD images is crucial in real-world applications such as referable DR screening. 
It allows us to notify about potential misclassifications to take appropriate actions in an informed way.
We proposed a  DPN-based referable DR screening framework that utilizes an OOD detector and a DR classifier to identify OOD images.
Experimental results on multiple real-world datasets demonstrate that incorporating a separate OOD detector can distinguish the OOD images, leading to decrease misclassification error.\\

\noindent\textbf{Acknowledgement.} This research is supported by the National
Research Foundation Singapore under its AI Singapore Programme
(AISG-GC-2019-001, AISG-RP-2018-008).

\bibliographystyle{IEEEbib}
\bibliography{retina}

\end{document}